\documentclass[letterpaper, 10 pt, conference]{ieeeconf}  

\IEEEoverridecommandlockouts                              

\usepackage{graphicx}
\usepackage{algorithm}
\usepackage{amsfonts} 
\let\labelindent\relax
\usepackage{enumitem}

\usepackage{wrapfig} 

\usepackage{amsmath}  
\usepackage{amssymb}  
\usepackage{bm}       
\usepackage{amsmath}
\usepackage{xcolor}
\usepackage{pifont}
\usepackage{etoolbox}
\usepackage{caption}
\usepackage{booktabs}
\usepackage{algorithm}
\usepackage{algorithmic}
\usepackage{pifont}
\usepackage{hyperref}
\title{\LARGE \bf
EasyInsert: A Data Efficient and Generalizable Insertion Policy
}

\author{Guanghe Li$^{1,2,3*}$, Junming Zhao$^{1,2,3*}$, Shengjie Wang$^{1,2,3}$, Yang Gao$^{1,2,3}$
\thanks{*Equal Contribution}
\thanks{$^{1}$Tsinghua University.}%
\thanks{$^{2}$Shanghai Qi Zhi Institute.}%
\thanks{$^{3}$Shanghai Artificial Intelligence Laboratory.}%
}

\begin{document}

\newcommand{\insertteaser}{
    \centering
    \vspace{5pt}
    {\small \href{https://easyinsert.github.io/}{\texttt{https://easyinsert.github.io/}}} \\
    \vspace{10pt}
    \includegraphics[width=1\textwidth]{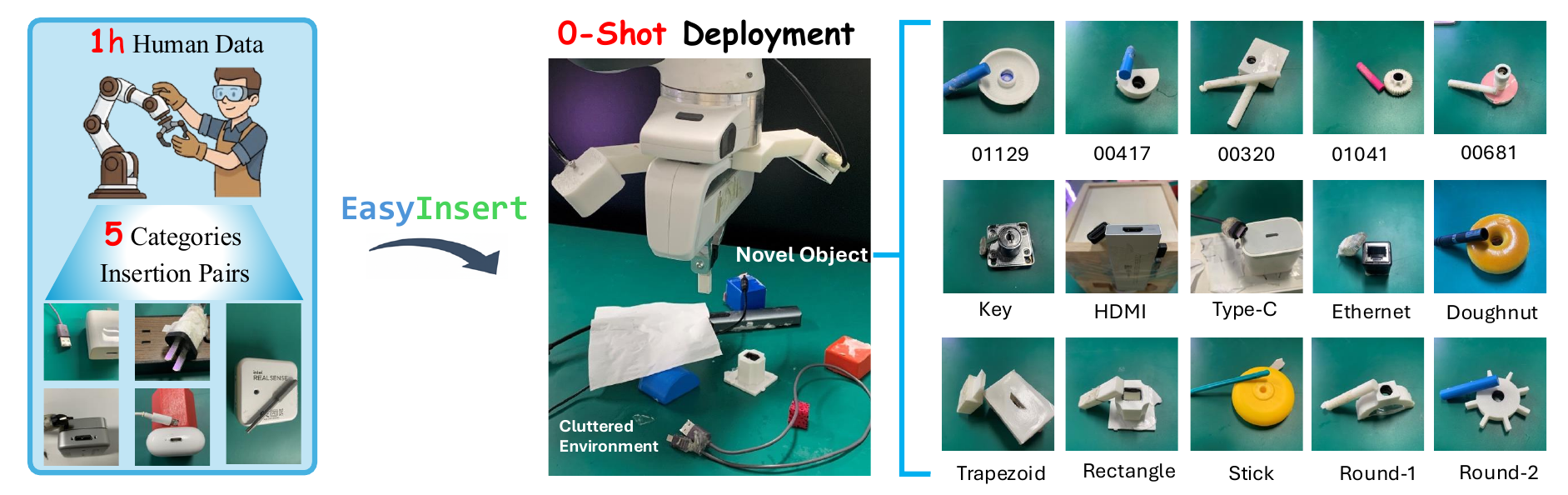}
    \captionof{figure}{EasyInsert is a highly efficient and generalizable robotic insertion framework. By leveraging just one hour of human teleoperation data to bootstrap a large-scale automated data collection process across five categories of insertion pairs, EasyInsert demonstrates strong zero-shot generalization to diverse, unseen objects. It remains robust even in densely cluttered environments and under significant initial pose offsets.}
    \label{fig:overall_fig1}
    \vspace{-10pt}
}

\makeatletter
\apptocmd{\@maketitle}{\insertteaser}{}{}
\makeatother

\maketitle

\setcounter{figure}{1}

\thispagestyle{empty}
\pagestyle{empty}


\begin{abstract}

Robotic insertion is a highly challenging task that requires exceptional precision in cluttered environments. Existing methods often have poor generalization capabilities. They typically function in restricted and structured environments, and frequently fail when the plug and socket are far apart, when the scene is densely cluttered, or when handling novel objects. They also rely on strong assumptions such as access to CAD models or a digital twin in simulation. To address these limitations, we propose EasyInsert. Inspired by human intuition, it formulates insertion as a delta-pose regression problem, which unlocks an efficient, highly scalable data collection pipeline with minimal human labor to train an end-to-end visual policy. During execution, the visual policy predicts the relative pose between plug and socket to drive a multi-phase, coarse-to-fine insertion process. EasyInsert demonstrates strong zero-shot generalization capability for unseen objects in cluttered environments, robustly handling cases with significant initial pose deviations. In real-world experiments, by leveraging just 1 hour of human teleoperation data to bootstrap a large-scale automated data collection process, EasyInsert achieves an over 90\% success rate in zero-shot insertion for 13 out of 15 unseen novel objects, including challenging objects like Type-C cables, HDMI cables, and Ethernet cables. Furthermore, requiring only a single manual reset, EasyInsert allows for fast adaptation to novel test objects through automated data collection and fine-tuning, achieving an over 90\% success rate across all 15 objects. 

\end{abstract}


\section{INTRODUCTION}

In modern factory assembly processes, workpieces of diverse shapes require precise assembly. To improve efficiency and achieve automation, some factories employ robots to replace human workers. However, conventional robotic assembly relies on deterministic control programs, necessitating extensive costs for customization. As a result, the pursuit of generalizable robotic assembly solutions has long garnered significant attention. In this work, we focus on insertion, a fundamental yet challenging skill in robotic assembly, characterized by contact-rich manipulation and high precision demands, placing higher demands on robot generalization.

Recent advances in imitation learning, reinforcement learning, and sim-to-real transfer have enabled robots to demonstrate robust performance across increasingly complex assembly tasks. Prior studies often import CAD models of specific workpieces into physics simulators, training agents via reinforcement learning before transferring them to real-world settings \cite{tang2024automate, tang2023industreal,narang2022factory,noseworthy2025forge}.  However, these approaches are heavily dependent on accurate CAD models and frequently suffer from performance degradation due to the sim-to-real gap. Moreover, while these works demonstrated their effectiveness on training artifacts, none of them tried to generalize to more unseen objects or diverse environments. Other works tried to overcome the sim2real gap by learning from real-world demonstration \cite{wen2022you, spector2022insertionnet,yu2023mimictouch} or applying RL directly in a real scenario \cite{luo2024serl, luo2024precise,xu2024rldg}. However, the high cost of collecting real-world data restricts these works to a small and clean environment, fitting models on limited training assembly pairs.  In summary, although these prior methods demonstrate competent performance under specific conditions, they fall short in generalizing to more realistic factory scenarios that demand: (1) 
 \textbf{Object generalization} – handling previously unseen objects with varied shapes and colors, (2) \textbf{Spatial generalization} – adapting to varied initial positions of the objects in a large space, (3) \textbf{Environmental generalization} – achieving robust performance across diverse environmental conditions.

In this work, we present \textbf{EasyInsert}, an efficient and generalizable framework for robotic insertion tasks. As shown in Figure \ref{fig:overall_fig1}, our generalist policy demonstrates strong generalization across diverse unseen objects, spatial configurations, and environmental conditions—without requiring additional efforts such as CAD model scanning or sim-to-real transfer.

We highlight the importance of sufficient, high-quality real-world data for achieving robust generalization in robotics. While collecting data in the real world is notorious for its high cost, we propose an efficient data collection framework that is both labor-light and scalable. Inspired by how humans perform insertions, we train a visual policy that predicts the delta position between plug and socket and introduce a coarse-to-fine execution strategy based on the prediction. In summary, our contributions include:

\begin{enumerate}

\item We propose \textbf{EasyInsert}, a framework that efficiently scales real-world training data and achieves strong generalization via delta-pose prediction followed by coarse-to-fine execution, handling unseen objects, diverse spatial configurations, and environmental variations. 

\item EasyInsert offers a series of advantages, (a) an automated data collection module that facilitates scalable and high-quality training data with minimal human effort, and (b) a generalist policy that generalizes effectively without requiring additional conditions such as precise CAD models or constrained workspaces. (c) an easy-to-use adaptation finetuning module that automatically collects data and achieves substantial improvements with only one environmental reset.

\item  EasyInsert demonstrates remarkable generalization capabilities, achieving over 90\% zero-shot success rates across a wide range of unseen objects, spatial positions, and even cluttered environments. When further enhanced with one-shot fine-tuning, it robustly maintains over 90\% success rates on all novel objects.
\end{enumerate}

\section{Related Work}
\label{sec:related_work}
We divide previous work on robotic insertion into two categories: Traditional methods and learning based methods. 


\textbf{Traditional Methods for Robotics Insertion.} Traditional approaches for robotic insertion generally rely on analytical modeling \cite{Whitney1982QuasiStaticAO, lozano1984automatic, mason2001mechanics, xia2006dynamic, newman2001interpretation} or heuristic-based search patterns \cite{haskiya1999robotic, bruyninckx1995peg, chhatpar2001search, park2013intuitive, sharma2013intelligent}. These traditional approaches  typically demand extremely high perception precision, require extensive manual parameter tuning, and are highly susceptible to modeling and sensing errors in real-world applications.

\textbf{Learning-based methods for Robotics Insertion.} In recent years, learning-based approaches have gained significant traction. One prominent paradigm leverages simulation environments for RL training followed by sim-to-real transfer \cite{tang2024automate, tang2023industreal, narang2022factory}. These approaches rely on the availability of accurate CAD models to build specific simulation environments for each object. Consequently, they often suffer from the sim-to-real gap and require extensive time to debug these environments and engineer effective reward functions. To improve sample efficiency and bridge these gaps, some approaches explore combinations of model-based planning and reinforcement learning \cite{marougkas2025integrating}, utilizing geometric or kinematic priors to guide policy exploration. Another line of work focuses on applying reinforcement learning or imitation learning directly in the real world \cite{zhao2022offline, learning_on_the_job, luo2024serl, luo2023rlif}, occasionally incorporating visual localization during deployment \cite{spector2022insertionnet}. Within this domain, human-in-the-loop \cite{luo2025precise} have been introduced to reduce the high cost of real-world data collection and interaction by leveraging human interventions to guide learning. Despite these efforts, such methods remain constrained by the need for continuous manual supervision and suffer from low data efficiency.

Ultimately, both paradigms exhibit significant limitations that hinder their deployment in realistic factory scenarios. Existing methods typically demand clean, structured workspaces and precise prior socket localization (often within 1–2 cm accuracy). Furthermore, they frequently struggle to generalize to unseen, out-of-distribution objects or require extensive, costly data collection. In contrast, our method is entirely CAD-free and highly data-efficient. EasyInsert achieves robust zero-shot insertion for novel objects, operates effectively in densely cluttered environments, and tolerates significant initial pose deviations (up to 6 cm in XY translation and $40^{\circ}$ in yaw orientation).



\begin{figure*}[ht]
    \centering
    \includegraphics[width=1\textwidth]{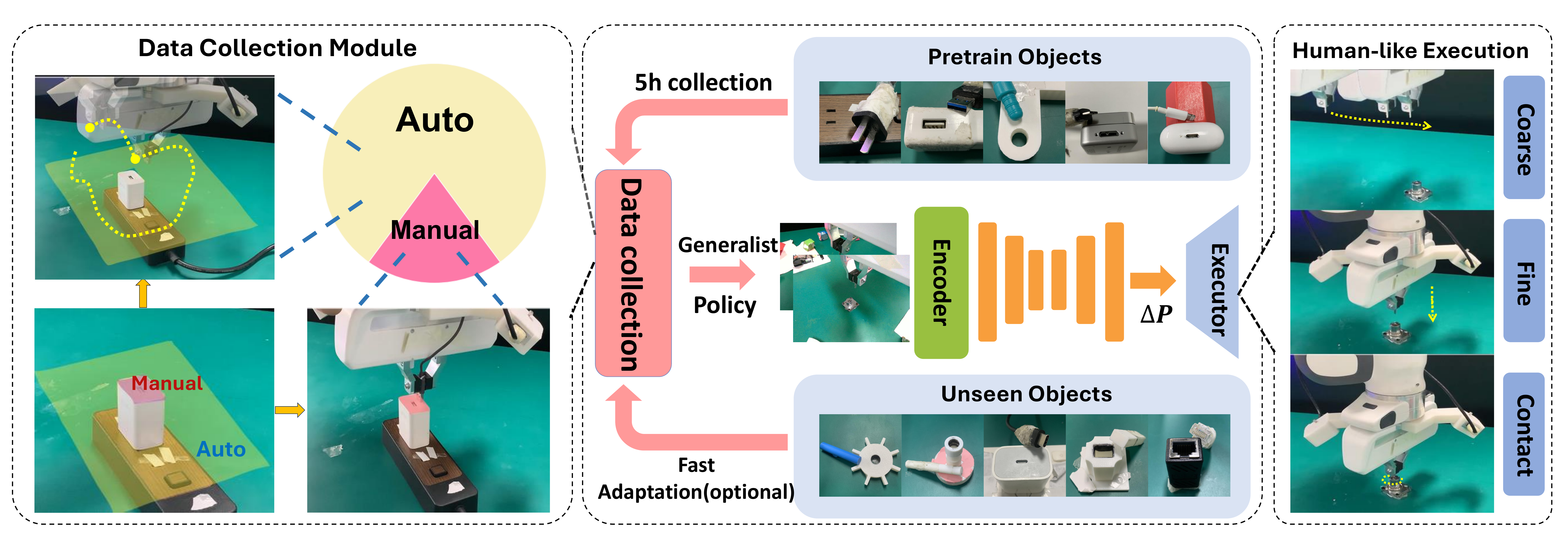}
    \caption{Overview of our method: (1) Left: Data collection module scaling data via 80\% automated spatial exploration and 20\% manual fine-grained interaction. (2) Middle: Generalist policy predicting relative pose directly from vision , supporting fast adaptation for novel test objects when requiring high precision. (3) Right: Coarse-to-fine execution motivated by human insertion behavior.} 
    \label{fig:method}
    \vspace{-10pt} 
\end{figure*}

\section{Method} 
\label{sec:method}

Inspired by how humans perform insertion, we designed our framework as 3 parts, shown in Figure \ref{fig:method}:  (1) \textbf{Spatially Decoupled Data Collection} that constructs training dataset through automated and manual data collection across 5 categories of objects, (2) \textbf{Generalist training} to predict relative pose between plug and socket directly from visual inputs, and (3) \textbf{Policy execution} that performs insertion using a coarse-to-fine strategy. When higher accuracy is required for novel test objects, we can fine-tune the generalist model by collecting additional data through our automated data collection module.

\subsection{Problem Formulation}
\label{sec:problem_formulation}

We define the robotic insertion task as the process of guiding a grasped plug to successfully mate with a target socket. Consistent with prior work \cite{tang2024automate, tang2023industreal}, we assume that in typical tabletop or industrial assembly-line setups, out-of-plane rotations (pitch and roll) are pre-constrained by the mechanical design of the end-effector fixtures or firmly established during the preceding structured grasp planning phase. Because the plug's orientation remains aligned with the robotic gripper throughout the task , we formulate this problem within a practical 4-Degree-of-Freedom (4-DoF) state space.

Let the current pose of the robot’s end-effector be denoted as $p=(x, y, z, \psi)$, where $(x, y, z)$ represents the 3D Cartesian position and $\psi$ represents the yaw angle. At any given time step $t$, the system receives visual observations $O_t = (O_{1,t}, O_{2,t})$ captured by dual wrist-mounted cameras. The primary objective is to maneuver the end-effector such that the plug is inserted into the socket.

\subsection{Key Insight: Insertion as Delta-Pose Regression}
\label{sec:key_insight}

\begin{figure}[b]
    \centering
    \vspace{-10pt} 
    \includegraphics[width=\columnwidth]
    {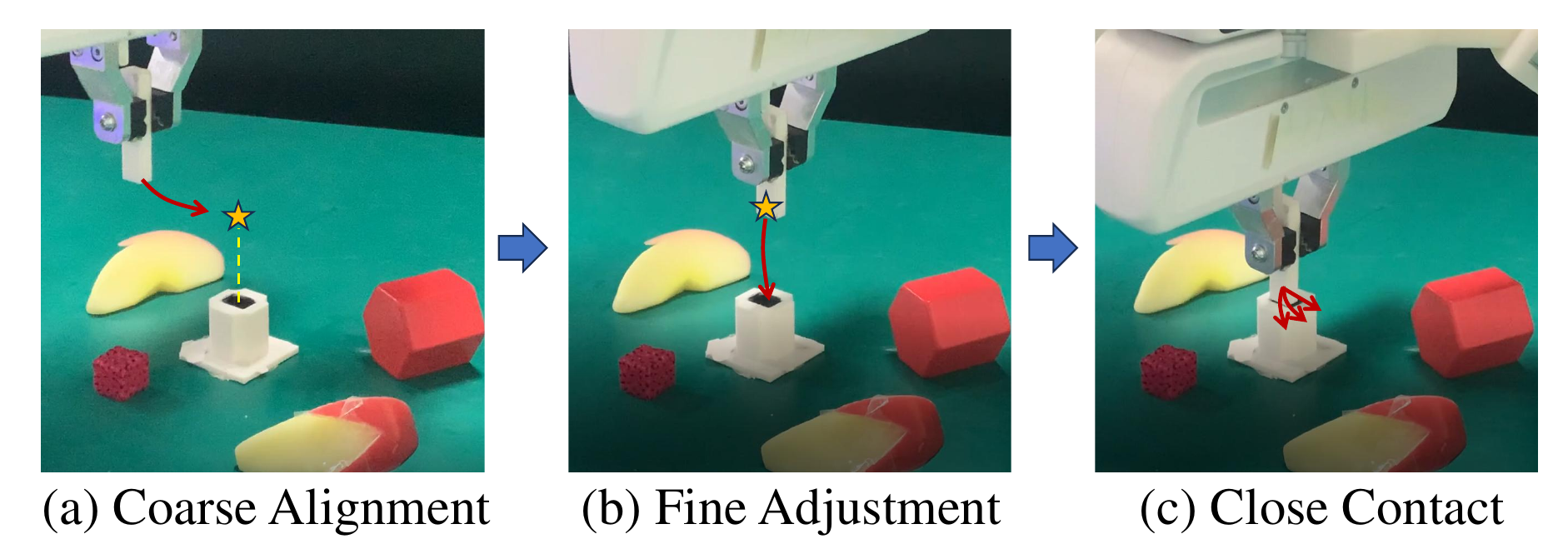}
    \caption{Coarse-to-fine hierarchical insertion procedure. } 
    \label{fig:three_step_demonstration}
\end{figure}

Previous learning-based robotic insertion methods have primarily focused on learning to predict precise robot actions from observations. However, these approaches face significant challenges, typically requiring extensive, high-quality human teleoperation or complex real-world reinforcement learning. In contrast, as shown in Figure \ref{fig:three_step_demonstration}, humans perform precise insertions by establishing an approximate alignment from a distance, refining it as they approach, and using subtle exploratory motions upon contact. The key to this strategy is continuously and roughly perceiving the relative pose (delta-pose) between the plug and the socket.

Inspired by this intuition, EasyInsert formulates the insertion task not as direct action prediction, but as a robust relative spatial regression problem. The overall pipeline operates as follows: during the training phase, our model learns from pre-collected interaction data  to predict the relative plug-socket pose $\Delta p = (\Delta x, \Delta y, \Delta z, \Delta \psi)$ from visual observations $(O_1, O_2)$ captured by two wrist-mounted cameras. During the testing and execution phase, instead of relying on the neural network to output high-frequency, low-level joint commands, this predicted delta-pose $\Delta p$ is fed into a multi-phase, closed-loop low-level controller (detailed in Section \ref{sec:execution}), which seamlessly translates the delta-pose into executable robot actions to complete the insertion. This paradigm offers two advantages:

\begin{itemize}
    \item{\textbf{Relaxed Data Quality Requirements: }}Unlike direct action prediction methods that fundamentally rely on continuous, expert-level trajectories to mitigate compounding errors, our formulation is entirely \textbf{agnostic to sequential trajectory quality}. Because the regression target $\Delta p$ depends solely on the deterministic geometric relationship between the gripper and the socket, every single collected data point is inherently valid and provides a precise ground-truth label. Consequently, EasyInsert can seamlessly learn from completely random spatial exploration, noisy interactions, or erratic movements, effectively eliminating the prerequisite for sequential expert demonstrations.

    \item{\textbf{Scalable, Automated Generation:}} Leveraging this relaxed data requirement, the system enables the autonomous generation of vast, diverse training datasets. By executing stochastic exploratory movements across a broad operational volume and instantly computing the corresponding relative offsets, \textbf{the dataset can be extensively scaled to encompass a wide-ranging distribution of relative poses}. This automated pipeline bypasses the bottleneck of continuous manual guidance, significantly reducing human labor while unlocking robust spatial generalization.
\end{itemize}

\subsection{Coarse-to-Fine Execution Strategy}
\label{sec:execution}

To effectively execute the predicted $\Delta p$, we introduce a multi-phase, closed-loop controller driven by real-time visual feedback. Rather than executing a static trajectory, the controller dynamically re-computes an absolute spatial waypoint $p_g$ at each timestep and directs the robot towards it. As shown in Figure \ref{fig:three_step_demonstration}, the controller continuously tracks the predicted $\Delta p$ to guide the end-effector through three distinct behavioral phases: 

\textbf{1) Coarse Alignment:} Initially, the controller prioritizes correcting large XY and rotational offsets. It aligns the plug with the socket in the $xy$-plane and adjusts the yaw angle $\psi$, while strictly maintaining a safe, vertical clearance above the socket to prevent collisions. 

\textbf{2) Fine Vertical Adjustment:} Once the horizontal and rotational alignments fall within a minimal error margin, the end-effector steadily lowers the plug along the $z$-axis. 

\textbf{3) Close-Contact with Perturbation:} During the final contact-rich phase, the controller introduces minor, random perturbations while continuing to apply downward pressure. This exploratory behavior effectively overcomes surface friction and naturally absorbs any residual micro-misalignments.

We provide the full details of the execution strategy in Algorithm \ref{alg:controller}.

\begin{algorithm}[H]
\caption{Coarse-to-Fine Execution Strategy}
\label{alg:controller}
\renewcommand{\algorithmiccomment}[1]{\hfill// #1}
\textbf{Input:} Initial gripper pose $p$, Policy $\pi$ \\
\textbf{Parameters:} Safety offset $H = 6$\,cm, Step size $d_z = 5$\,mm \\
\textbf{Output:} Successful insertion
\begin{algorithmic}[1]
\WHILE{task not completed}
    \STATE Capture visual observations $(O_1, O_2)$
    \STATE // Predict delta-pose
    \STATE $\Delta p = (\Delta x, \Delta y, \Delta z, \Delta \psi) \leftarrow \pi(O_1, O_2)$ 

    \STATE $g \leftarrow p + \Delta p$ 
    
    \IF{$\sqrt{\Delta x^2 + \Delta y^2} > 2\text{cm}$ \OR $|\Delta \psi| > 20^\circ$ \OR $\Delta z > 1\text{cm}$} 
        \STATE  \textbf{// Phase 1: Coarse Alignment}
        \STATE $p_{\text{next}} \leftarrow (g_x, g_y, g_z + H, g_{\psi})$
        
    \ELSIF{$g_z + 1\text{cm} < p_z$}
        \STATE \textbf{// Phase 2: Fine Vertical Adjustment}
        \STATE $p_{\text{next}} \leftarrow (g_x, g_y, p_z - d_z, g_{\psi})$ 
        
    \ELSE
        \STATE \textbf{// Phase 3: Close-Contact with Perturbation}
        \STATE $\text{noise} \sim \mathcal{U}(-3\text{mm}, 3\text{mm})$
        \STATE $p_{\text{next}} \leftarrow (g_x + \text{noise}, g_y + \text{noise}, p_z - d_z, g_{\psi})$
    \ENDIF
    
    \STATE $\text{Execute\_Motion}(p_{\text{next}})$
    \STATE $p \leftarrow \text{Get\_Current\_Pose}()$
\ENDWHILE
\end{algorithmic}
\end{algorithm}


\subsection{Spatially Decoupled Data Collection}

Building on the relaxed data requirements discussed in Section \ref{sec:problem_formulation}, we implement a spatially decoupled data collection pipeline directly in the real world. This native real-world approach allows us to effortlessly incorporate everyday objects and natural physical distractors, generating diverse, cluttered environments while eliminating both the laborious asset modeling required in simulators and the sim2real gap during deployment.

Ultimately, the goal of this pipeline is to construct a training dataset $\mathcal{D} = \{ (O_{1}, O_{2}, \Delta p) \}$. To achieve this efficiently, we first record the gripper's end-effector pose as the target pose $p_{goal}$, when the plug is perfectly placed into the socket. During data collection, for any current gripper pose $p_{current}$, the ground-truth delta-pose is instantly computed via a relative coordinate transformation:
\begin{equation}
    \Delta p = p_{goal} \ominus p_{current}
\end{equation}
This straightforward geometric calculation automatically provides the precise $\Delta p$ labels paired with the corresponding visual observations $O_{1}$ and $O_{2}$, serving as the key enabler for our highly scalable, hybrid data collection pipeline.

\textbf{1) Automated Free-Space Collection (-8 cm to +8 cm):} For the vast majority of the workspace, data collection is entirely automated. A motion planner rapidly samples and navigates the gripper to random, collision-free poses within a large bounding box around the socket (up to 8 cm in XY translation and $40^{\circ}$ in yaw). As the gripper moves, the system continuously captures image pairs $(O_{1}, O_{2})$ and records the corresponding $\Delta p$ at 10-15 Hz. During this automated process, a human supervisor only needs to occasionally introduce visual diversity by randomly tossing or rearranging distracting objects in the background. As evaluated in Section \ref{sec:exp_ablation}, a policy trained exclusively on automated data achieves robust coarse alignment and vertical adjustment. Notably, without a single human demonstration, it navigates the plug to within 1 cm of the socket with near 100\% success and even yields a meaningful zero-shot success rate on novel objects. However, the system frequently fails during the final millimeter-level close-contact phase. This specific limitation strongly motivates our subsequent targeted manual collection.

\textbf{2) Manual Close-Contact Collection (-1 cm to +1 cm):} To bridge the performance gap in the final millimeters of insertion, we incorporate manual data collection. While automated sampling near the socket is possible, random exploration at such close proximity risks unsafe collisions. Instead, an operator uses kinesthetic teaching to safely guide the robot within a confined region of -1 cm to +1 cm. This targeted intervention demands minimal human effort while safely securing high-quality data for the final insertion phase. Crucially, this kinesthetic teaching does not require the operator to perform a perfect, continuous insertion trajectory. Instead, the operator simply introduces random, noisy spatial perturbations (e.g., erratic wiggling) while maintaining close contact. Because our model performs regression on relative spatial states, this unstructured, noisy exploration is not only sufficient but actually highly effective for densely covering the state distribution near the socket.

By spatially decoupling the workspace, EasyInsert achieves exceptional data efficiency. Of the 5 hours of training data collected across 5 categories, 80\% is generated autonomously. Human effort is minimized to just 1 hour of close-contact manual demonstrations and minor environment randomization.

\subsection{Model Training}

\textbf{Model Architecture and Pretraining.} Given the dataset $\mathcal{D}=\{(O^i_1, O^i_2, \Delta p^i)\}_{i=1}^{N}$, we parameterize our $\Delta p$ prediction model using a Diffusion Policy \cite{chi2023diffusion} that takes dual wrist-camera images $(O_1, O_2)$ as input. Despite the target pose being deterministic, Diffusion Policy provides superior stability and expressiveness for high-dimensional visual-to-spatial regression compared to standard MSE estimators. Specifically, we extract visual features from the images using a shared ResNet18 \cite{he2016deep} encoder pre-trained on ImageNet. These features are then fused and fed into the diffusion action head for relative pose prediction. To enable fast, real-time inference during our closed-loop execution, we utilize Denoising Diffusion Implicit Models (DDIM) as our sampling accelerator. Furthermore, to ensure robustness against lighting variations and distractors, we apply aggressive color augmentations (e.g., strong jittering, random grayscale conversion) during training. This forces the model to learn geometry-centric representations rather than overfitting to specific colors or backgrounds. 

 
\textbf{Rapid Adaptation via Target Registration. }Although the base policy exhibits strong zero-shot performance, out-of-distribution objects with tight tolerances (such as Type-C ports) may require enhanced precision. EasyInsert supports a rapid adaptation phase that requires minimal human intervention. An operator simply mates the plug with the socket to record the final target pose, $p_{goal}$. Following this setup, the system autonomously utilizes the free-space collection module to generate approximately 4 minutes of fine-tuning data. Costing only 30 seconds of human effort, this target-conditioned process efficiently bridges the precision gap for challenging geometries.

\begin{table*}[t]
\vspace*{5pt}

\centering
\caption{Comparison with baseline algorithms on AutoMate \cite{tang2024automate} Objects. While all baseline methods strictly rely on exact CAD models, \textbf{EasyInsert achieves superior performance without requiring any CAD access.}}
\label{tab:table_1_comparison}
\resizebox{0.8\linewidth}{!}{
\begin{tabular}{lccccccc}
\toprule
\textbf{Method} & \textbf{01129} & \textbf{00417} & \textbf{00320} & \textbf{01041} & \textbf{00681} &\textbf{OOD} &\textbf{CAD-Free}\\
\midrule
AutoMate-s & \textbf{10/10} & \textbf{10/10} & 8/10 & 6/10 & 6/10 & \ding{55} & \ding{55} \\
AutoMate-g & 7/10 & 9/10 & \textbf{10/10} & 6/10 & 5/10 & \ding{55} & \ding{55} \\
FP-Insert & 0/10 & 1/10 & 2/10 & 0/10 & 0/10 & $\checkmark$ & \ding{55} \\
EasyInsert & \textbf{10/10} & \textbf{10/10} & \textbf{10/10} & \textbf{9/10} & \textbf{9/10} & $\checkmark$ & $\checkmark$\\
\bottomrule
\end{tabular}
}
\end{table*}

\section{Experiments}
\label{sec:experiments}

We present a comprehensive evaluation of EasyInsert through 15 real-world insertion tasks, along with ablation studies validating its key design choices.

\subsection{Setup} \label{sec:exp_setup}







\begin{figure}[b] 
    \centering
    \vspace{-10pt}
    \includegraphics[width=\columnwidth]{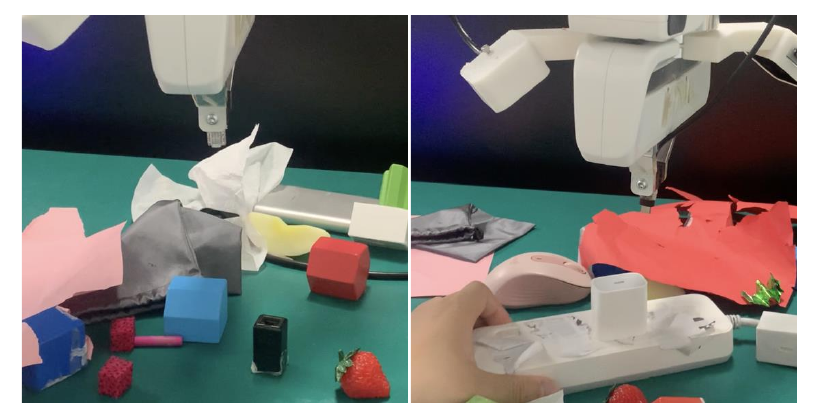}
    \caption{Left: We randomly place distraction objects around the socket. Right: In perturbation experiments, we randomly move socket positions.}
    \label{fig:distract_display}
\end{figure}

\textbf{Hardware Setup.}  As shown in Figure \ref{fig:overall_fig1}, the robotic system consists of a 7-DoF Franka Emika Panda arm, using dual wrist-mounted Intel Realsense 405 RGB cameras for visual perception.  

\textbf{Task definitions.} We trained EasyInsert on five categories of objects and evaluated it on 15 novel insertion tasks as illustrated in Figure \ref{fig:overall_fig1}. During evaluation, the socket position was randomized within the workspace, while the plug's initial pose varied with XY-plane offsets of 3–6 cm and yaw angles of 15–40 degrees relative to the socket. To assess environmental robustness, we introduced random distraction objects near the socket, as demonstrated in Figure \ref{fig:distract_display}.

\subsection{Baselines} \label{sec:baselines}
We evaluate our method against the following baselines:

\textbf{AutoMate:} AutoMate \cite{tang2024automate} constructs a simulation environment for each plug-socket pair utilizing ground-truth CAD models. It engineers reward functions to train an insertion policy via PPO and DAgger, followed by sim-to-real transfer for real-world deployment. We evaluate two variants: \textbf{AutoMate-s}, which represents the individual specialist policies trained for specific objects, and \textbf{AutoMate-g}, which distills these specialist policies into a unified generalist model.

\textbf{FP-Insert:} We introduce a baseline combining FoundationPose \cite{wen2024foundationpose} and motion planning. Assuming access to CAD models, FoundationPose accurately estimates object poses. FP-Insert leverages these estimates to compute the relative pose between the plug and the socket, subsequently executing the insertion via a motion planner.

Note that all baseline methods strictly require precise CAD models , and the AutoMate variants are restricted to in-domain objects encountered during training. To the best of our knowledge, EasyInsert is the first approach to successfully achieve zero-shot insertion of unseen objects under challenging conditions where the plug and socket are initialized with significant spatial separation.

\begin{table}[t]
\centering
\caption{Zero-shot evaluation on 10 unseen objects. Unlike the objects in Table~\ref{tab:table_1_comparison}, \textbf{these objects have no available CAD files}, demonstrating the method's generalization capability to novel geometries.}
\label{tab:table_2_comparison}
\resizebox{\linewidth}{!}{
\begin{tabular}{lccccc}
\toprule
\textbf{Object} & \textbf{Key} & \textbf{HDMI} & \textbf{TypeC} & \textbf{Ethernet} & \textbf{Doughnut} \\
\midrule
EasyInsert & 10/10 & 9/10 & 8/10 & 9/10 & 10/10 \\
\midrule
\textbf{Object} & \textbf{Trapezoid} & \textbf{Rectangle} & \textbf{Stick} & \textbf{Round-1} & \textbf{Round-2} \\
\midrule
EasyInsert & 8/10 & 10/10 & 10/10 & 9/10 & 9/10 \\
\bottomrule
\end{tabular}
}
\vspace{-10pt}
\end{table}

\subsection{Overall Results} \label{sec:exp_zeroshot} 

Trained on only 1 hour of human data and 5 hours of data in total, EasyInsert exhibited strong zero-shot generalization across all 15 unseen objects, achieving success rates of over 90\% on most tasks as shown in Tables \ref{tab:table_1_comparison} and \ref{tab:table_2_comparison}.

\textbf{Comparison to baselines.} As shown in Table \ref{tab:table_1_comparison}, EasyInsert demonstrates superior performance despite operating out-of-distribution (OOD) and without CAD models. In contrast, the AutoMate variants achieve lower success rates even when evaluated in-domain with precise CAD models. Furthermore, FP-Insert fails entirely across all tasks. This failure is twofold: (1) the FoundationPose delta-pose estimation is inherently noisy, and (2) its standard motion planner is unsuited for contact-rich insertion. Instead of aligning the plug prior to insertion, the planner executes a direct, end-to-end trajectory. Compounded by estimation errors, this approach causes the plug to frequently collide with the socket's exterior walls, become stuck on the surface, or miss the hole entirely. When transitioning to novel OOD objects without CAD files (Table \ref{tab:table_2_comparison}), all baseline algorithms experience a complete collapse in performance (0\% success rate), leaving EasyInsert as the only successful method.


\textbf{Object Generalization.} Despite training on only five categories, EasyInsert demonstrates remarkable generalization to unseen shapes. This is driven by efficient, workspace-spanning data collection and delta-pose prediction, which relies on consistent relative spatial relationships. Consequently, the system effectively handles novel shapes (e.g., HDMI), color variations (e.g., Ethernet) , and underrepresented geometries like Trapezoid. The slightly lower 80\% success rate for Type-C and trapezoidal plugs is expected, as their extremely tight socket tolerances demand exceptionally high precision.

\textbf{Spatial Generalization.}  
EasyInsert demonstrates robust spatial generalization, effectively handling the significant initial pose deviations defined in our setup. To further validate its limits, we conducted additional rollouts in the HDMI environment with extreme out-of-distribution perturbations (10 cm offset in XY), achieving a 3/5 success rate. Notably, even with coarse initial pose estimates in these extreme OOD scenarios, the system effectively guides the gripper towards the socket, performing successful insertions.

\textbf{Environment Generalization and Robustness.} EasyInsert demonstrates strong resilience to both static and dynamic environmental variations. During evaluation, the presence of distractor objects around the socket did not degrade performance, a robustness we attribute to the manual introduction of environmental perturbations during the automated data collection phase. To further evaluate the system's stability, we conducted dynamic perturbation experiments by randomly altering the positions of the socket and surrounding distractors during execution. As shown in Table \ref{tab:purturbation}, EasyInsert maintained highly consistent performance despite these active disruptions. This stability is fundamentally enabled by our closed-loop execution strategy, which continuously adjusts to ensure that sudden positional changes do not affect the final insertion outcome

\begin{table}[htbp] 
    \centering
    \begin{tabular}{lccc} 
        \toprule
        \textbf{Object Name} & \textbf{Rectangle} & \textbf{Ethernet} & \textbf{Type-C} \\  
        \midrule 
        EasyInsert & 5/5 & 5/5 & 4/5 \\    
        \bottomrule
    \end{tabular}   
    \caption{Perturbation experiment.} 
    \label{tab:purturbation}
    \vspace{-10pt}
\end{table}

\subsection{Automated Fine-Tuning for Challenging Geometries} \label{sec:exp_oneshot}
For novel objects requiring higher insertion precision, we conducted rapid fine-tuning using in-domain data gathered via our automated collection module. With only 30 seconds of manual human effort for initial target registration, the system autonomously generated 4 minutes of targeted training data. After 10 minutes of fine-tuning the policy on a single RTX 3090 GPU, we achieved a 100\% success rate on all previously failing objects, as shown in Table \ref{tab:finetuned}. This demonstrates highly efficient spatial adaptation, completing the full data collection and fine-tuning pipeline in under 15 minutes.

\begin{table}[htbp] 
    \centering
    \begin{tabular}{|c|c|c|} 
        \hline
        \rule{0pt}{2.0ex}
        \textbf{Object Name} & \textbf{Zero-shot} & \textbf{Fine-tuned}  \\
        \hline
        \rule{0pt}{2.0ex}
        Trapezoid & 8/10 & \textbf{10/10}  \\
        Type-C & 8/10 & \textbf{10/10} \\
        \hline
    \end{tabular}
    \caption{Fine-Tuning Performance.}
    \label{tab:finetuned}
    \vspace{-10pt}
\end{table}

\subsection{Ablation Study} \label{sec:exp_ablation} 

In this section, we examine the key designs of EasyInsert across three insertion environments: Type-C, Ethernet, and Stick.

\begin{figure}[htbp] 
    \centering
    \includegraphics[width=\linewidth]{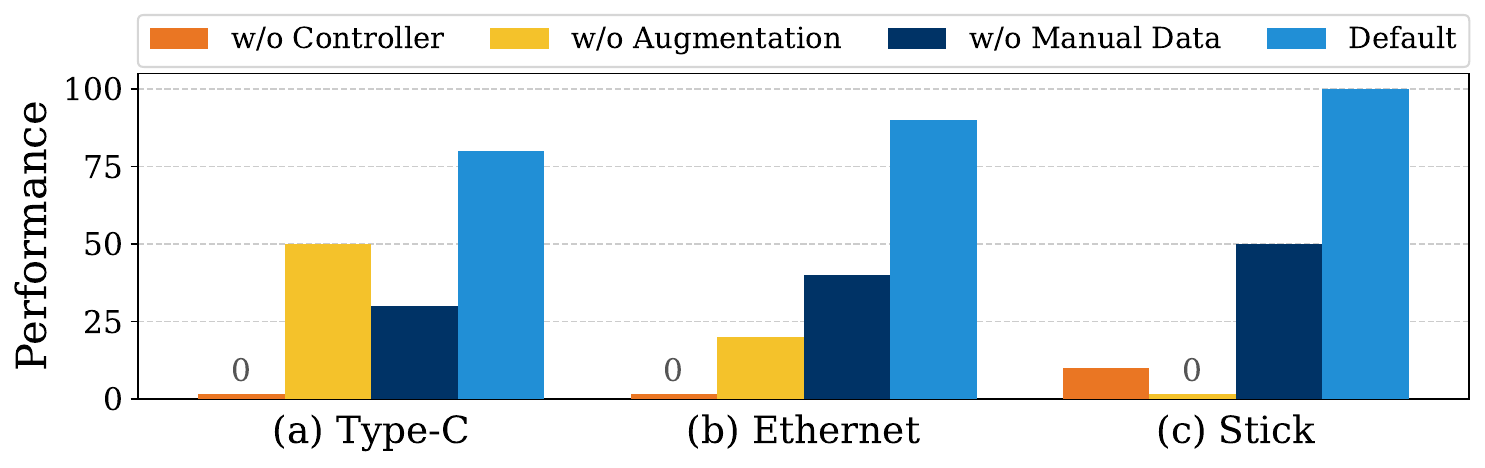}
    \caption{Ablation study on different components of EasyInsert}
    \label{fig:manual_abl}
\end{figure}

\textbf{Ablation Study on data augmentation.}  We investigate the role of data augmentation during pretraining. As shown in Figure \ref{fig:manual_abl}, removing image augmentations caused a 67\% performance drop, particularly for colors that are rare in training data(stick) and transparent objects like Ethernet. This shows the importance of augmentation in bridging domain gaps between training and novel objects.

\textbf{Ablation Study on Coarse-to-fine Controller. }
We evaluate the necessity of our multi-phase execution strategy by replacing it with a standard motion planner that moves directly to the predicted target pose. As shown in Figure \ref{fig:manual_abl}, executing a direct, end-to-end trajectory without prior spatial alignment leads to the exact same failure modes (e.g., collisions and surface sticking) observed in the FP-Insert baseline. This confirms that standard motion-planners are fundamentally unsuited for contact-rich insertion.

\textbf{Ablation Study on Data Collection Strategy.} EasyInsert pretrains on a hybrid dataset comprising both automated and manually collected data. To evaluate the necessity of each component, we train and compare two separate policies: \textbf{\texttt{w/o Manual Data}} (automated data only) and \textbf{\texttt{w/o Automated Data}} (manual data only).

First, to address whether automated data is truly necessary, we evaluated the \textbf{\texttt{w/o Automated Data}} policy on the three insertion environments. Because our manual data is exclusively collected within a highly confined proximity (-1 cm to +1 cm), deploying this policy in our standard evaluation setting with initial spatial deviations of 3-6 cm results in a 0\% success rate(thus omitted from Figure \ref{fig:manual_abl}). The large initial offsets are entirely out-of-distribution (OOD) for the manual-only policy, causing the robot to exhibit erratic movements and fail before reaching the socket. This validates that relying solely on manual collection is strictly insufficient for spatial generalization.

\begin{table}[h]
\centering

\begin{tabular}{lccc}
\hline
\textbf{Object} & \textbf{Final Success} & \textbf{$<$ 1 cm} & \textbf{$<$ 5 mm} \\
\hline
Type-C   & \phantom{0}3 / 10 & 10 / 10 & \phantom{0}6 / 10 \\
Ethernet & \phantom{0}4 / 10 &  \phantom{0}9 / 10 & \phantom{0}7 / 10 \\
Stick    & \phantom{0}5 / 10 & 10 / 10 & \phantom{0}8 / 10 \\
\hline
\end{tabular}
\caption{Success Rates at Different Proximity Thresholds (No Manual Data)}
\label{tab:proximity_ablation}
\end{table}

\begin{figure}[htbp]
    \centering
    \includegraphics[width=\linewidth]{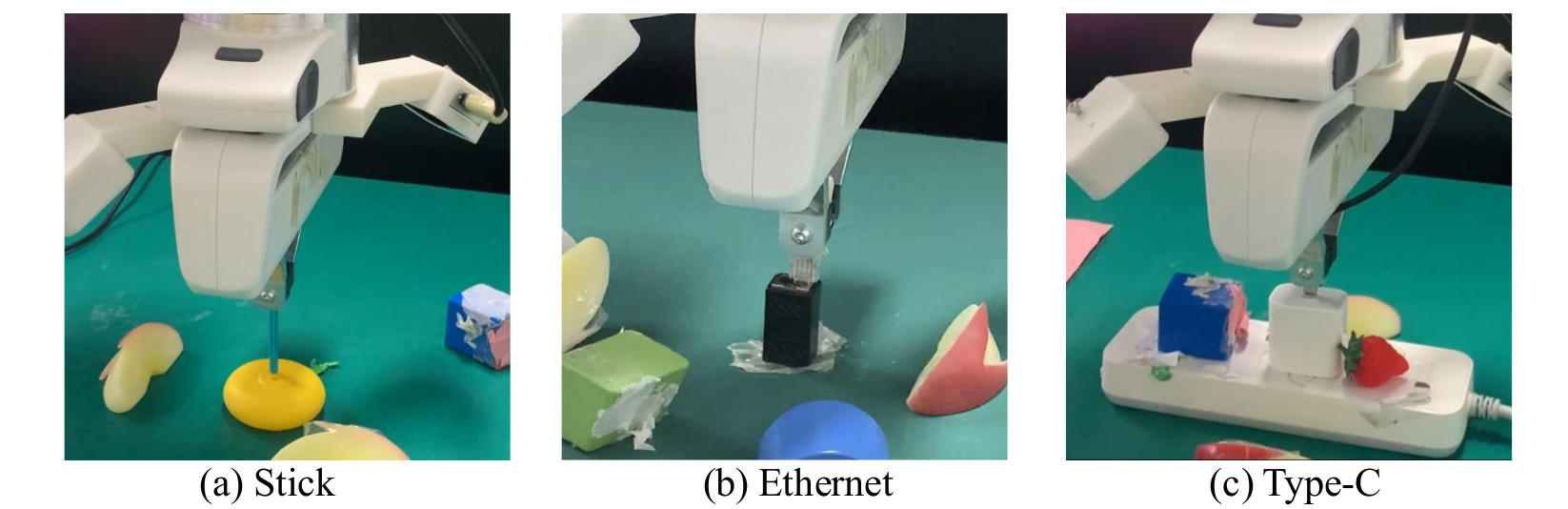}
    \caption{Visualization of failure cases without manual data.}
    \label{fig:all_failure_case_display}
\end{figure}

Conversely, we evaluated the \textbf{\texttt{w/o Manual Data}} policy to understand the limits of automated collection. While excluding manual data leads to an overall decline in final insertion success (Figure \ref{fig:manual_abl}), the \textbf{\texttt{w/o Manual Data}} policy remarkably maintains a near 100\% success rate in guiding the plug to within 1 cm of the socket (Table \ref{tab:proximity_ablation}). However, performance drops significantly at the 5 mm threshold. Failure case analysis (Figure \ref{fig:all_failure_case_display}) confirms that the automated-only system excels at coarse alignment and fine vertical adjustment but struggles during the final close-contact phase.

Ultimately, these results demonstrate that both data sources are indispensable and highly complementary. Automated data efficiently solves the large-scale spatial alignment problem, while targeted manual data is strictly required for mastering the final high-precision, millimeter-level contact.

\begin{figure}[htbp] 
    \centering
    \vspace{5pt}
    \includegraphics[width=\linewidth]{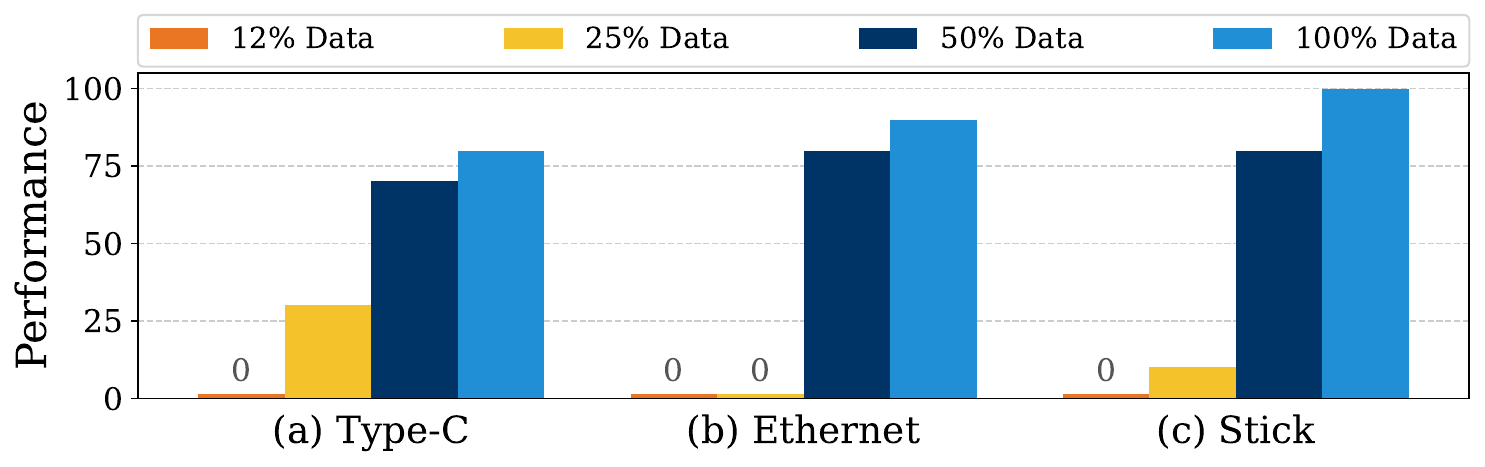}
    \caption{Ablation study on data amount}
    \label{fig:amount_abl}
    \vspace{-10pt}
\end{figure}

\textbf{Ablation Study on Data Amount. }  We study the impact of pretraining data amount. Experimental results shown in Figure \ref{fig:amount_abl} illustrate that more pre-training data provides better generalization and performance. Notably, trained on 50\% of data, EasyInsert maintains 76\% zero-shot success rate. However, when the training data is reduced to 12\%, the system fails to generalize to novel objects, highlighting the importance of sufficient training samples for robust performance.

\section{Conclusion}
\label{sec:conclusion}
In this paper, we present EasyInsert, an efficient and generalizable robotic insertion framework. Inspired by human intuition, we frame insertion as a delta-pose regression task, enabling a scalable data collection pipeline driven by just one hour of human teleoperation to train an end-to-end visual policy. Coupled with a coarse-to-fine execution strategy, this policy empowers EasyInsert to achieve an over 90\% zero-shot success rate on 13 novel objects. Extensive experiments demonstrate its robustness in densely cluttered environments and under significant initial pose deviations. Furthermore, the system supports rapid, automated fine-tuning with minimal human effort to master high-precision novel tasks. We believe this framework establishes a robust foundation for advancing general-purpose robotic assembly.

\bibliographystyle{IEEEtran} 
\bibliography{reference}          

\section{Appendix}

\subsection{Train Object Display}
\begin{figure}[H]
    \centering
    \vspace{-10pt}
    \includegraphics[width=\linewidth]{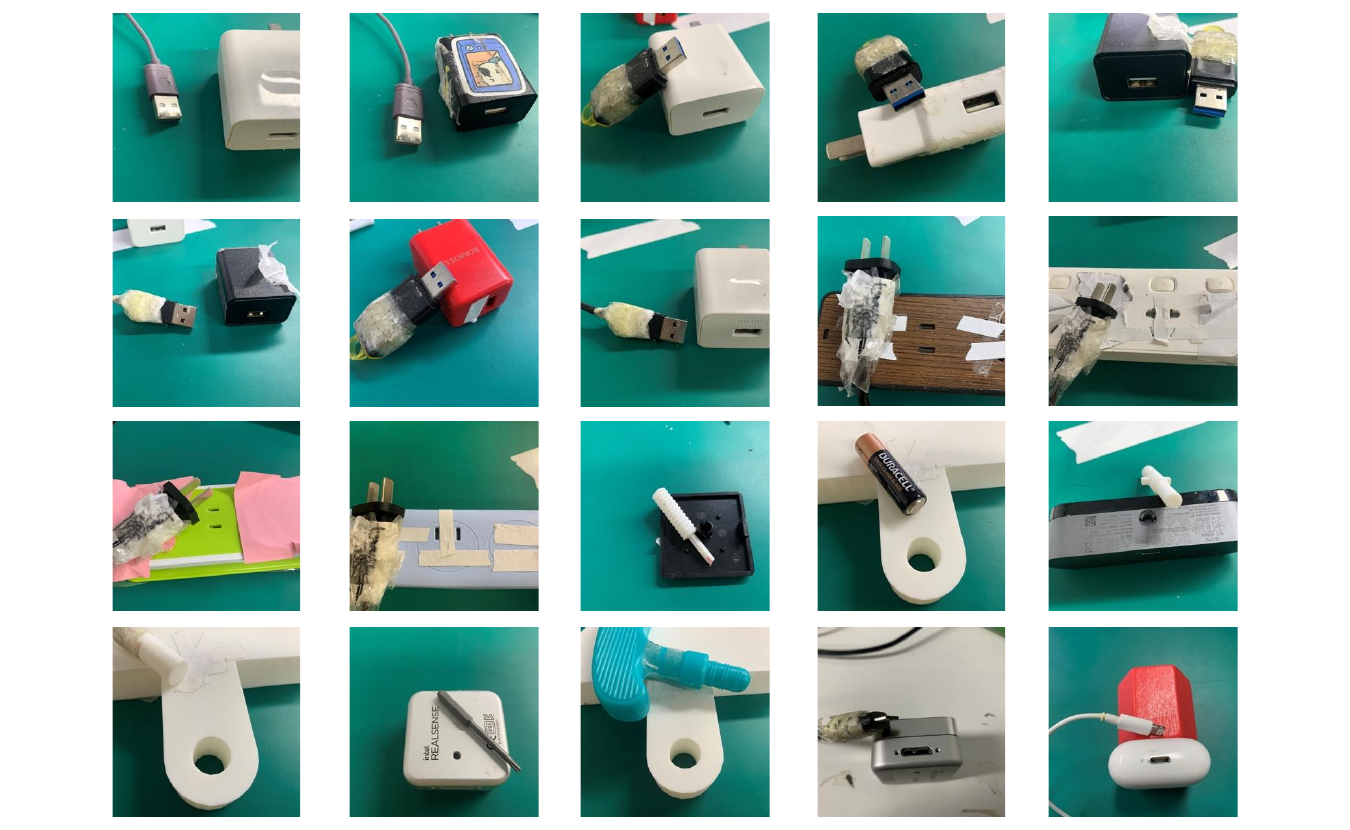}
    \caption{Visualization of training objects.}
    \label{fig:train_object_display}
    \vspace{-10pt}
\end{figure}

In Figure \ref{fig:train_object_display}, we visualize all the training objects used in our study. EasyInsert is trained on five categories of common objects, including: 8 pairs of USB connectors, 4 pairs of plugs, 6 pairs of round objects,1 MicroUSB 3.0 connector and 1 Apple charger. 
Notably, all selected objects are everyday items that can be easily obtained in household or laboratory environments.

\label{sec:appendix_train_object_display}

\subsection{Video Frames of EasyInsert}

\begin{figure}[H] 
    \centering
    \includegraphics[width=\linewidth]{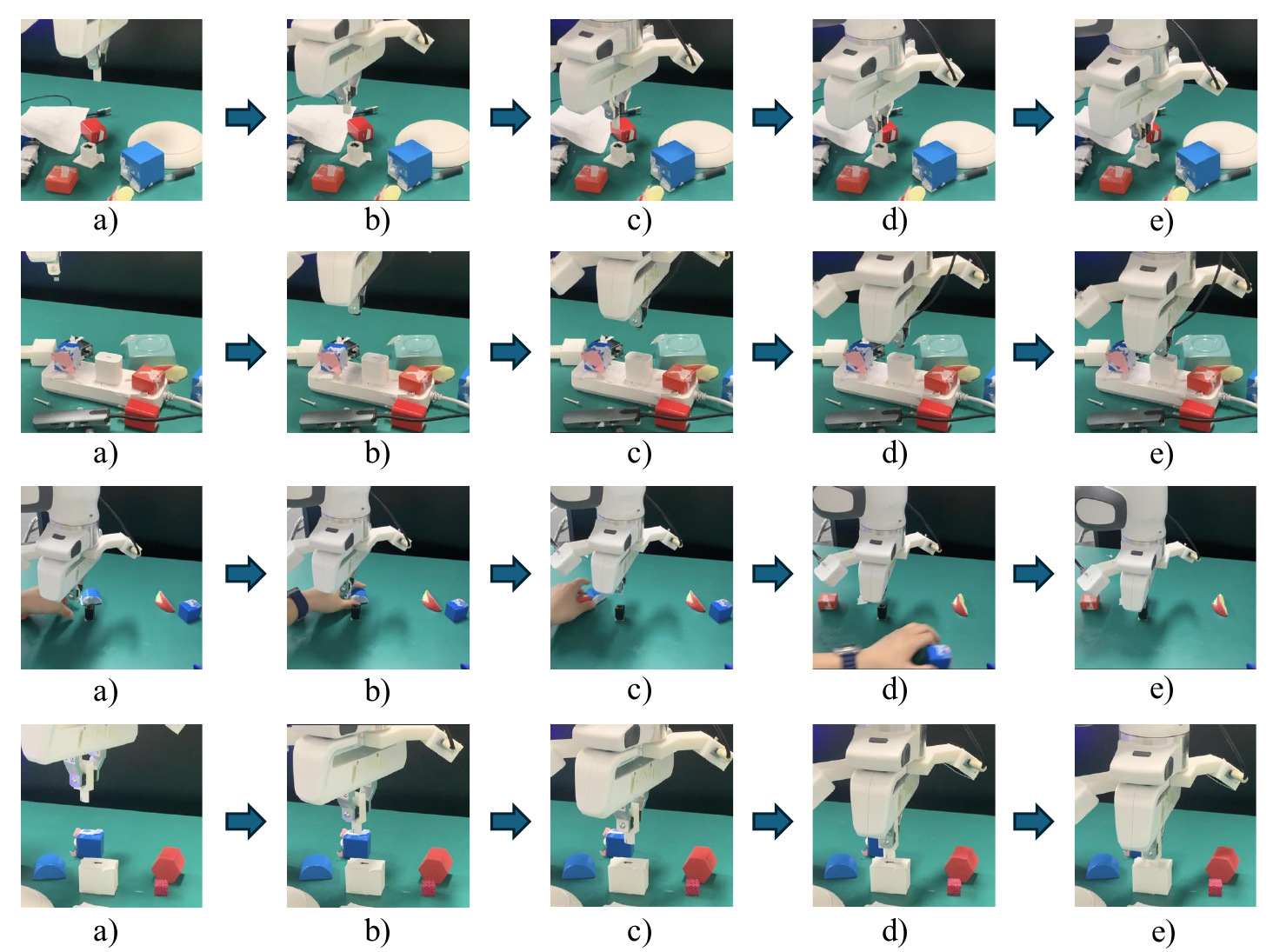}
    \caption{Video frames of EasyInsert inserting novel objects in cluttered environments(Please refer to the supplementary video file for the full dynamic process). }
    \label{fig:video_snap_display}
    \vspace{-10pt}
\end{figure}

We here present video snaps of EasyInsert inserting different objects (Rectangle, Type-C, Ethernet and Trapezoid) in Figure \ref{fig:video_snap_display}. We note that during evaluation, we place random distraction objects around the socket. In third row during perturbation experiments, we randomly move socket position and surrounding objects to examine the robustness of EasyInsert. As demonstrated in the video clips, EasyInsert achieves successful zero-shot insertion even in cluttered environments and with large initial pose deviations. Please refer to the supplementary video file for the full dynamic process.

\end{document}